\def\year{2018}\relax
\documentclass[letterpaper]{article} 
\usepackage{aaai18}  
\usepackage{times}  
\usepackage{helvet}  
\usepackage{courier}  
\usepackage{url}  
\usepackage{graphicx}  
\usepackage[utf8]{inputenc} 
\usepackage[T1]{fontenc}    
\usepackage{hyperref}       
\usepackage{url}            
\usepackage{booktabs}       
\usepackage{amsfonts}       
\usepackage{nicefrac}       
\usepackage{microtype}      
\usepackage{graphicx}
\usepackage{color}
\usepackage{algorithm}
\usepackage{algorithmic}
\usepackage{amsmath,amssymb,amsthm}\usepackage[toc,page]{appendix}
\usepackage[font=small,labelfont=bf]{caption}
\newcommand{\x}{\textbf{x}}
\newcommand{\y}{\textbf{y}}
\newcommand{\z}{\textbf{z}}
\newcommand{\g}{\textbf{g}}
\newcommand{\h}{\textbf{h}}
\newcommand{\sleec}{\textsc{Sleec}}
\newcommand{\leml}{\textsc{Leml}}
\newcommand{\dxml}{\textsc{DXML}}
\newcommand{\pdsparse}{\textsc{Pd-Sparse}}
\newcommand{\dismec}{\textsc{DiSMEC}}
\newcommand{\fastxml}{\textsc{FastXML}}
\newcommand{\pfastrexml}{\textsc{PfastreXML}}
\newcommand{\lemlimc}{\textsc{Leml-Imc}}
\newcommand{\onevsall}{\textsc{One-vs-All}}
\newcommand{\exmldsone}{\textsc{ExMLDS1}}
\newcommand{\exmldstwo}{\textsc{ExMLDS2}}
\newcommand{\wsabie}{\textsc{Wsabie}}
\newcommand{\exmldsthree}{\textsc{ExMLDS3}}
\newcommand{\pmi}{\textsf{PMI}}
\newcommand{\ppmi}{\textsf{PPMI}}
\newcommand{\sppmi}{\textsf{SPPMI}}
\newcommand{\annexml}{\textsc{AnnexML}}
\newcommand{\ppdsparse}{\textsc{PPDsparse}}

\usepackage{blindtext}

\title{Leveraging Distributional Semantics for \\ Multi-Label Learning}
\author{Rahul Wadbude \\ IIT Kanpur \\ \texttt{warahul@iitk.ac.in}  \And Vivek Gupta \\ Microsoft Research \\ \texttt{t-vigu@microsoft.com} \And Piyush Rai \\ IIT Kanpur \\ \texttt{piyush@iitk.ac.in} \AND Nagarajan Natarajan \\ Microsoft Research \\ \texttt{t-nanata@microsoft.com} \And Harish Karnick \\ IIT Kanpur \\ \texttt{hk@iitk.ac.in} \And Prateek Jain \\ Microsoft Research \\ \texttt{prajain@microsoft.com} }

\begin{document}


\maketitle

\begin{abstract}
We present a novel and scalable label embedding framework for large-scale multi-label learning a.k.a ExMLDS (Extreme Multi-Label Learning using Distributional Semantics). Our approach draws inspiration from ideas rooted in distributional semantics, specifically the Skip Gram Negative Sampling (SGNS) approach, widely used to learn word embeddings for natural language processing tasks. Learning such embeddings can be reduced to a certain matrix factorization. Our approach is novel in that it highlights interesting connections between label embedding methods used for multi-label learning and paragraph/document embedding methods commonly used for learning representations of text data. The framework can also be easily extended to incorporate auxiliary information such as label-label correlations; this is crucial especially when there are a lot of missing labels in the training data. We demonstrate the effectiveness of our approach through an extensive set of experiments on a variety of benchmark datasets, and show that the proposed learning methods perform favorably compared to several baselines and state-of-the-art methods for large-scale multi-label learning. To facilitate end-to-end learning, we develop a joint learning algorithm that can learn the embeddings as well as a regression model that predicts these embeddings given input features, via efficient gradient based methods. 
\end{abstract}

\section{Introduction}
Modern data generated in various domains are increasingly "multi-label" in nature; images (e.g. \emph{Instagram}) and documents (e.g. \emph{Wikipedia}) are often identified with multiple tags, online advertisers often associate multiple search keywords with ads, and so on. Multi-label learning is the problem of learning to assign multiple labels to instances, and has received a great deal of attention over the last few years; especially so, in the context of learning with millions of labels, now popularly known as extreme multi-label learning ~\cite{jain2016extreme,bhatia2015sparse,babbar2016dismec,prabhu2014fastxml}. 

The key challenges in multi-label learning, especially when there are millions of labels, include a) the data may have a large fraction of labels missing, and b) the labels are often heavy-tailed~\cite{bhatia2015sparse,jain2016extreme} and predicting labels in the tail becomes significantly hard for lack of training data. For these reasons, and the sheer scale of data, traditional multi-label classifiers are rendered impracticable. State-of-the-art approaches to extreme multi-label learning fall broadly under two classes: 1) embedding based methods, e.g.~\leml~\cite{yu2014large}, \wsabie ~\cite{weston2010large}, \sleec~ \cite{bhatia2015sparse}, \pdsparse~ \cite{yen2016pd}), and 2) tree-based methods~\cite{prabhu2014fastxml,jain2016extreme}. The first class of approaches are generally scalable and work by embedding the high-dimensional label vectors to a lower-dimensional space and learning a regressor in that space. In most cases, these methods rely on a key assumption that the binary label matrix is low rank and consequently the label vectors can be embedded into a lower-dimensional space. At the time of prediction, a decompression matrix is used to retrieve the original label vector from the low-dimensional embeddings. As corroborated by recent empirical evidence~\cite{bhatia2015sparse,jain2016extreme}, approaches based on standard structural assumptions such as low-rank label matrix fail and perform poorly on the tail. The second class of methods (tree-based) methods for multi-label learning try to move away from rigid structural assumptions~\cite{prabhu2014fastxml,jain2016extreme}, and have been demonstrated to work very well especially on the tail labels. 

In this work, we propose an embedding based approach, closely following the framework of ~\sleec~\cite{bhatia2015sparse}, that leverages a word vector embedding technique~\cite{mikolov2013distributed} which has found resounding success in natural language processing tasks. Unlike other embedding based methods, \sleec{} has the ability to learn \emph{non-linear} embeddings by aiming to preserve only local structures and example neighborhoods. We show that by learning rich \texttt{word2vec} style embedding for instances (and labels), we can a) achieve competitive multi-label prediction accuracies, and often improve over the performance of the state-of-the-art embedding approach~\sleec{} and b) cope with missing labels, by incorporating auxiliary information in the form of label-label co-occurrences, which most of the state-of-the-art methods can not. Furthermore, our learning algorithm admits significantly faster implementation compared to other embedding based approaches. The distinguishing aspect of our work is that it draws inspiration from distributional semantics approaches~\cite{mikolov2013distributed,le2014distributed}, widely used for learning non-linear representations of text data for natural language processing tasks such as understand word and document semantics, classifying documents, etc. 

Our main contributions are:
\begin{enumerate}
\item We leverage an interesting connection between the problem of learning distributional semantics in text data analysis and the multi-label learning problem. To the best of our knowledge, this is a novel application.
\item The proposed objectives for learning embeddings can be solved efficiently and scalably; the learning reduces to a certain matrix factorization problem.
\item Unlike existing multi-label learning methods, our method can also leverage label co-occurrence information while learning the embeddings; this is especially appealing when a large fraction of labels are missing in the label matrix.
\item We show improvement in training time as compared to state-of-art label embedding methods for extreme multi-label learning, while being competitive in terms of label prediction accuracies; we demonstrate scalability and prediction performance on several state-of-the-art moderate-to-large scale multi-label benchmark datasets.
\end{enumerate}

The outline of the paper is as follows. We begin by setting up notation, background and describing the problem formulation in Section~\ref{sec:probform}. In Section~\ref{sec:wordvec}, we present our training algorithms based on learning word embeddings for understanding word and document semantics. Here we propose two objectives, where we progressively incorporate auxiliary information \emph{viz.} label correlations. We present comprehensive experimental evaluation in Section~\ref{sec:experiments}, and conclude.

\section{Problem Formulation and Background}
\label{sec:probform}
In the standard multi-label learning formulation, the learning algorithm is given a set of training instances $\{\x_1, \x_2, \dots, \x_n \}$, where $\x_i \in \mathbb{R}^d$ and the associated label vectors $\{\y_1, \y_2, \dots, \y_n\}$, where $\y_i \in \{0,1\}^L$. In real-world multi-label learning data sets, one does not usually observe irrelevant labels; here $y_{ij} = 1$ indicates that the $j$th label is \emph{relevant} for instance $i$ but $y_{ij} = 0$ indicates that the label is \emph{missing} or \emph{irrelevant}. Let $Y \in \{0,1\}^{n \times L}$ denote the matrix of label vectors. In addition, we may have access to label-label co-occurrence information, denoted by $C \in \mathbb{Z}_+^{L \times L}$ (e.g., number of times a pair of labels co-occur in some external source such as the Wikipedia corpus). The goal in multi-label learning is to learn a vector-valued function $\mathbf{f}: \x \mapsto \mathbf{s}$, where $\mathbf{s} \in \mathbb{R}^L$ scores the labels. 

Embedding-based approaches typically model $\mathbf{f}$ as a composite function $\mathbf{h}(\mathbf{g}(\x))$ where, $\mathbf{g}: \mathbb{R}^d \to \mathbb{R}^{d'}$ and $\mathbf{h}:\mathbb{R}^{d'} \to \mathbb{R}^L$. For example, assuming both $\mathbf{g}$ and $\mathbf{h}$ as linear transformations, one obtains the formulation proposed by~\cite{yu2014large}. The functions $\mathbf{g}$ and $\mathbf{h}$ can be learnt using training instances or label vectors, or both. More recently, non-linear embedding methods have been shown to help improve multi-label prediction accuracies significantly. In this work, we follow the framework of \cite{bhatia2015sparse}, where $\mathbf{g}$ is a linear transformation, but $\mathbf{h}$ is non-linear, and in particular, based on $k$-nearest neighbors in the embedded feature space.

In \sleec{}, the function $\g: \mathbb{R}^d \to \mathbb{R}^{d'}$ is given by $\g(\x) = V\x$ where $V \in \mathbb{R}^{d' \times d}$. The function $\h: \mathbb{R}^{d'} \to \mathbb{R}^L$ is defined as:

\begin{equation} 
\h \bigg(\mathbf{z}; \{\z_i, \y_i\}_{i=1}^n\bigg) = \frac{1}{|\mathcal{N}_k|}\sum_{i \in \mathcal{N}_k} \y_i, 
\label{eqn:knn}
\end{equation}

where $\z_{i} = \g(\x_i)$ and $\mathcal{N}_k$ denotes the $k-$nearest neighbor training instances of $\z$ in the embedded space. Our algorithm for predicting the labels of a new instance is identical to that of \sleec{} and is presented for convenience in Algorithm~\ref{alg:pred}. Note that, for speeding up predictions, the algorithm relies on clustering the training instances $\x_i$; for each cluster of instances $Q^\tau$, a different linear embedding $\mathbf{g}_\tau$, denoted by $V^\tau$, is learnt.

\begin{algorithm}
\caption{Prediction Algorithm}
\label{alg:pred}
\begin{algorithmic}                   
\STATE \textbf{Input}: Test point: $\x$, no. of nearest neighbors $k$, no. of desired
labels $p$.
\STATE 1. $Q_\tau$ : partition closest to $\x$.
\STATE 2. $\z \leftarrow V^{\tau}\x$.
\STATE 3. $\mathcal{N}_k \leftarrow k$ nearest neighbors of $\z$ in the embedded instances of $Q_\tau$.
\STATE 4. $\mathbf{s} = \h(\z; \{\z_i, \y_i \}_{i\in Q_\tau})$ where $\h$ is defined in \eqref{eqn:knn}.
\STATE \textbf{return} top $p$ scoring labels according to $\mathbf{s}$.
\end{algorithmic}
\end{algorithm}

In this work, we focus on learning algorithms for the functions $\g$ and $\h$, inspired by their successes in natural language processing in the context of learning distributional semantics~\cite{mikolov2013distributed,levy2014neural}. In particular, we use techniques for inferring word-vector embeddings for learning the function $\h$ using a) training label vectors $\y_i$, and b) label-label correlations $C \in \mathbb{R}^{L \times L}$. 

Word embeddings are desired in natural language processing in order to understand semantic relationships between words, classifying text documents, etc. Given a text corpus consisting of a collection of documents, the goal is to embed each word in some space such that words appearing in similar \emph{contexts} (i.e. adjacenct words in documents) should be \emph{closer} in the space, than those that do not. In particular, we use the \texttt{word2vec} embedding approach~\cite{mikolov2013distributed} to learn an embedding of instances, using their label vectors $\y_1, \y_2, \dots, \y_n$. \sleec{} also uses nearest neighbors in the space of label vectors $\y_i$ in order to learn the embeddings. However, we show in experiments that \texttt{word2vec} based embeddings are richer and help improve the prediction performance significantly, especially when there is a lot of missing labels. In the subsequent section, we discuss our algorithms for learning the embeddings and the training phase of multi-label learning.

\section{Learning Instance and Label Embeddings}
\label{sec:wordvec}
There are multiple algorithms in the literature for learning word embeddings~\cite{mikolov2013distributed,pennington2014glove}. In this work, we use the Skip Gram Negative Sampling~(SGNS) technique, for two reasons a) it is shown to be competitive in natural language processing tasks, and more importantly b) it presents a unique advantage in terms of scalability, which we will address shortly after discussing the technique.

\paragraph{Skip Gram Negative Sampling.} In SGNS, the goal is to learn an embedding $\z \in \mathbb{R}^{d'}$ for each word $w$ in the vocabulary. To do so, words are considered in the \emph{contexts} in which they occur; context $c$ is typically defined as a fixed size window of words around an occurrence of the word. The goal is to learn $\z$ such that the words in similar contexts are closer to each other in the embedded space. Let $w' \in c$ denote a word in the context $c$ of word $w$. Then, the likelihood of observing the pair $(w,w')$ in the data is modeled as a sigmoid of their inner product similarity:
\[ P( \text{Observing } (w,w') ) = \sigma(\langle \z_{w}, \z_{w'} \rangle) = \frac{1}{1 + \exp(\langle-\z_w, \z_{w'}\rangle)} \ .\]
To promote dissimilar words to be further apart, negative sampling is used, wherein randomly sampled negative examples $(w, w'')$ are used. Overall objective favors $\z_w, \z_{w'}, \z_{w''}$ that maximize the log likelihood of observing $(w.w')$, for $w' \in c$, and the log likelihood of $P( \text{not observing } (w,w'') ) = 1 - P( \text{Observing } (w,w'') )$ for randomly sampled negative instances. Typically, $n_-$ negative examples are sampled per observed example, and the resulting SGNS objective is given by:
\begin{equation}
\begin{split}
 \max_{\z} \sum_{w} \bigg( {\sum_{w': (w', w)}} \log\big(\sigma( \langle \z_{w}, \z_{w'} \rangle)\big) + \\ \frac{n_-}{\#w} {\sum_{w''}} \log\big(\sigma(-  \langle \z_{w}, \z_{w{''}} \rangle )\big)\bigg),   
 \label{eqn:sgns}
 \end{split}
\end{equation}
where $\#w$ denotes the total number of words in the vocabulary, and the negative instances are sampled uniformly over the vocabulary.

\subsection{Embedding label vectors}
\label{subsec:labelvec}
We now derive the analogous embedding technique for multi-label learning. A simple model is to treat each instance as a "word"; define the "context" as $k$-nearest neighbors of a given instance in the space formed by the training label vectors $\y_i$, with cosine similarity as the metric. We then arrive at an objective identical to \eqref{eqn:sgns} for learning embeddings $\z_1, \z_2, \dots, \z_n$ for instances $\x_1, \x_2, \dots, \x_n$ respectively:

\begin{equation}
\small
\begin{split}
 \max_{\z_1,\z_2,\dots,\z_n} \sum_{i=1}^n \bigg( {\sum_{j: \mathcal{N}_k(\y_i)}} \log\big(\sigma( \langle \z_{i}, \z_{j} \rangle)\big) + \\ \frac{n_-}{n} \sum_{j'} \log\big(\sigma(-  \langle \z_{i}, \z_{j'} \rangle )\big)\bigg),   
 \label{eqn:sgnsmll}
 \end{split}
\end{equation}
Note that $\mathcal{N}_k(\y_i)$ denotes the $k$-nearest neighborhood of $i$th instance in the space of \emph{label vectors}~\footnote{Alternately, one can consider the neighborhood in the $d$-dimensional feature space $\x_i$; however, we perform clustering in this space for speed up, and therefore the label vectors are likely to preserve more discriminative information within clusters.} or instance embedding.
After learning label embeddings $\z_i$, we can learn the function $\g: \x \to \z$ by regressing $\x$ onto $\z$, as in \sleec. Solving~\eqref{eqn:sgnsmll} for $\z_i$ using standard \texttt{word2vec} implementations can be computationally expensive, as it requires training multiple-layer neural networks. Fortunately, the learning can be significantly sped up using the key observation by~\cite{levy2014neural}.

\cite{levy2014neural} showed that solving SGNS objective is equivalent to matrix factorization of the \emph{shifted positive point-wise mutual information} (SPPMI) matrix defined as follows. Let $M_{ij} = \langle \y_i, \y_j \rangle$.
\[
\pmi_{ij}(M) = \log\left(\frac{M_{ij}*|M|}{\sum_k {M_{(i,k)}}*\sum_k{M_{(k,j)}}}\right)
\]
\begin{equation}
\sppmi_{ij}(M) = \max(\pmi_{ij}(M) - \log(k),0)
\label{eqn:sppmi}
\end{equation}
Here, \pmi{} is the point-wise mutual information matrix of $M$ and $|M|$ denotes the sum of all elements in $M$. Solving the problem \eqref{eqn:sgnsmll} reduces to factorizing the shifted PPMI matrix $M$. 

Finally, we use ADMM ~\cite{boyd2011distributed} to learn the regressors $V$ over the embedding space formed by $\z_i$. Overall training algorithm is presented in \ref{alg:PPMI}.
\begin{algorithm}
\caption{Learning embeddings via SPPMI factorization~(\exmldsone{}).}
\label{alg:PPMI}
\begin{algorithmic}                   
\STATE \textbf{Input.} Training data $(\x_i, \y_i), i = 1,2,\dots,n$.

\STATE 1. Compute $\widehat{M} := \sppmi(M)$ in \eqref{eqn:sppmi}, where $M_{ij} = \langle \y_i, \y_j \rangle$.

\STATE 2. Let $U,S,V = \text{svd}(\widehat{M})$, and preserve top $d'$ singular values and singular vectors.

\STATE 3. Compute the embedding matrix $Z = U{S}^{0.5}$, where $Z \in \mathbb{R}^{n \times d'}$, where $i$th row gives $\z_i$

\STATE 4. Learn $V$ s.t. $XV^T = Z$ using ADMM~\cite{boyd2011distributed}, where $X$ is the matrix with $\x_i$ as rows.
\STATE \textbf{return} V, Z
\end{algorithmic}
\end{algorithm}

We refer to Algorithm \ref{alg:PPMI} based on fast \ppmi{} matrix factorization for learning label vector embeddings as~\exmldsone{}. We can also optimize the objective \ref{eqn:sgnsmll} using a neural network model~\cite{mikolov2013distributed}; we refer to this \texttt{word2vec} method for learning embeddings in Algorithm~\ref{alg:PPMI} as~\exmldstwo{}.

\subsection{Using label correlations}
\label{subsec:labelcorr}
In various practical natural language processing applications, superior performance is obtained using joint models for learning embeddings of text documents as well as individual words in a corpus~\cite{dai2015document}. For example, in PV-DBoW~\cite{dai2015document}, the objective while learning embeddings is to maximize similarity between embedded documents and words that compose the documents. Negative sampling is also included, where the objective is to minimize the similarity between the document embeddings and the embeddings of high frequency words. In multi-label learning, we want to learn the embeddings of labels as well as instances jointly. Here, we think of labels as individual words, whereas label vectors (or instances with the corresponding label vectors) as paragraphs or documents. As alluded to in the beginning of Section~\ref{sec:wordvec}, in many real world problems, we may also have auxiliary label correlation information, such as label-label co-occurrence. We can easily incorporate such information in the joint modeling approach outlined above. To this end, we propose the following objective that incorporates information from both label vectors as well as label correlations matrix: 

\begin{equation}
  \max_{z,\bar{z}} \mathbb{O}_{z,\bar{z}} =  \mu_1{\mathbb{O}^1_{\bar{z}}} +  \mu_2{\mathbb{O}^2_{z}} + 
 \mu_3{\mathbb{O}^3}_{\{z,\bar{z}\}}
  \label{eqn:sgnsmllcomb}
\end{equation}


\begin{equation}
\small
\begin{split}
\mathbb{O}^1_{\bar{z}} = \sum_{i=1}^L \bigg( {\sum_{j: \mathcal{N}_k(C(i,:))}} \log\big(\sigma( \langle \bar{\z}_{i}, \bar{\z}_{j} \rangle)\big) + \\ \frac{n^1_-}{L} \sum_{j'} \log\big(\sigma(-  \langle \bar{\z}_{i}, \bar{\z}_{j'} \rangle )\big)\bigg),   
 \label{eqn:sgnssmll1}
\end{split}
\end{equation}


\begin{equation}
\small
\begin{split}
\mathbb{O}^2_{z} = \sum_{i=1}^n \bigg( {\sum_{j: \mathcal{N}_k(M(i,:))}} \log\big(\sigma( \langle \z_{i}, \z_{j} \rangle)\big) + \\ \frac{n^2_-}{n} \sum_{j'} \log\big(\sigma(-  \langle \z_{i}, \z_{j'} \rangle )\big)\bigg),   
 \label{eqn:sgnsmll2}
\end{split}
\end{equation}


\begin{equation}
\small
\begin{split}
 \mathbb{O}^3_{\{z,\bar{z}\}} = \sum_{i=1}^L \bigg( {\sum_{j: y_{ij} = 1}} \log\big(\sigma( \langle \z_{i}, \bar{\z}_{j} \rangle)\big) + \\ \frac{n^3_-}{L} \sum_{j'} \log\big(\sigma(-  \langle \z_{i}, \bar{\z}_{j'} \rangle )\big)\bigg)  
 \label{eqn:sgnsmll3}
\end{split}
\end{equation}

Here, $\z_i$, $i = 1,2,\dots,n$ denote embeddings of instances while $\bar{\z}_{i}$, $i = 1,2,\dots,L$ denote embeddings of labels. $\mathcal{N}_k(M(i,:))$ denotes the $k$-nearest neighborhood of $i$th instance in the space of label vectors. $\mathcal{N}_k(C(i,:))$ denotes the $k$-nearest neighborhood of $i$th label in the space of \emph{labels}. Here, $M$ defines instance-instance correlation i.e. $M_{ij} = \langle \y_i, \y_j \rangle$ and $C$ is the label-label correlation matrix. Clearly, \eqref{eqn:sgnsmll2} above is identical to \eqref{eqn:sgnsmll}. $\mathbb{O}^1_{\bar{\z}}$ tries to embed labels $\bar{\z}_{i}$ in a vector space, where correlated labels are closer; $\mathbb{O}^2_{\z}$ tries to embed instances $\z_i$ in such a vector space, where correlated instances are closer; and finally, $\mathbb{O}^3_{\{\z,\bar{\z}\}}$ tries to embed labels and instances in a common space where labels occurring in the $i$th instance are closer to the embedded instance.

Overall the combined objective $\mathbb{O}_{\{\z,\bar{\z}\}}$ promotes learning a common embedding space where correlated labels, correlated instances and observed labels for a given instance occur closely. Here $\mu_1$,$\mu_2$ and $\mu_3$ are hyper-parameters to weight the contributions from  each type of correlation. $n^1_-$ negative examples are sampled per observed label, $n^2_-$ negative examples are sampled per observed instance in context of labels and $n^3_-$ negative examples are sampled per observed instance in context of instances. Hence, the proposed objective efficiently utilizes label-label correlations to help improve embedding and, importantly, to cope with missing labels. The complete training procedure using ~\sppmi{} factorization is presented in Algorithm \ref{alg:PPMI2}. Note that we can use the same arguments given by ~\cite{levy2014neural} to show that the proposed combined objective~\eqref{eqn:sgnsmllcomb} is solved by ~\sppmi{} factorization of the joint matrix $A$ given in Step 1 of Algorithm~\ref{alg:PPMI2}. 

\begin{algorithm}
\caption{Learning joint label and instance embeddings via SPPMI factorization~(\exmldsthree{}).}
\label{alg:PPMI2}
\begin{algorithmic}                   
\STATE \textbf{Input.} Training data $(\x_i, \y_i), i = 1,2,\dots,n$ and C (label-label correlation matrix) and objective weighting $\mu_1$,$\mu_2$ and $\mu_3$.

\STATE 1. Compute $\widehat{A} := \sppmi(A)$ in \eqref{eqn:sppmi}; write \[ A = 
\begin{pmatrix}
\mu_2M & \mu_3Y \\
\mu_3Y^T & \mu_1C\\
\end{pmatrix},
\] $M_{ij} = \langle \y_i, \y_j \rangle$, Y is label matrix with $\y_i$ as rows.

\STATE 2. Let $U,S,V = \text{svd}(\widehat{A})$, and preserve top $d'$ singular values and singular vectors.

\STATE 3. Compute the embedding matrix $Z = U{S}^{0.5}$; write \[ Z = 
\begin{pmatrix}
Z_1 \\
Z_2\\
\end{pmatrix},
\] where rows of $Z_1 \in \mathbb{R}^{n \times d'}$ give instance embedding and rows of $Z_2 \in \mathbb{R}^{L \times d'}$ give label embedding.

\STATE 4. Learn $V$ s.t. $XV^T = Z_1$ using ADMM~\cite{boyd2011distributed}, where $X$ is the matrix with $\x_i$ as rows. 
\STATE \textbf{return} V, Z
\end{algorithmic}
\end{algorithm}

\begin{algorithm}[h!]
\caption{Prediction Algorithm with Label Correlations ~(\exmldsthree{} prediction).}
\begin{algorithmic}                 
\STATE \textbf{Input}: Test point: $\x$, no. of nearest neighbors $k$, no. of desired labels $p$, $V$, embeddings $Z_1$ and $Z_2$.
\STATE 1. Use Algorithm \ref{alg:pred} (Step 3) with input $Z_1,k,p$ to get score $s_1$. 
\STATE 3. Get score $s_2 = Z_2Vx$
\STATE 4. Get final score $s$ = $\frac{s_1}{\|s_1\|}$ + $\frac{s_2}{\|s_2\|}$.
\STATE \textbf{return} top $p$ scoring labels according to $\mathbf{s}$.
\end{algorithmic}
\label{alg:pred2}
\end{algorithm}

At test time, given a new data point we could use the Algorithm \ref{alg:pred} to get top $p$ labels. Alternately, we propose to use Algorithm \ref{alg:pred2} that also incorporates similarity with label embeddings $Z_2$ along with $Z_1$ during prediction, especially when there are very few training labels to learn from. In practice, we find this prediction approach useful. Note the $\z_i$ corresponds to the $i^{th}$ row of $Z_1$, and $\bar{\z}_j$ corresponds to the $j^{th}$ row of $Z_2$. We refer the Algorithm \ref{alg:PPMI2} based on the combined learning objective \eqref{eqn:sgnsmllcomb} as~\exmldsthree{}.

\section{Experiments}
\label{sec:experiments}
We conduct experiments on commonly used benchmark datasets from the extreme multi-label classification repository provided by the authors of~\cite{prabhu2014fastxml,bhatia2015sparse} \footnote{http://manikvarma.org/downloads/XC/XMLRepository.html}; these datasets are pre-processed, and have prescribed train-test splits. Statistics of the datasets used in experiments is shown in Table \ref{tab:datasets}. We use the standard, practically relevant, precision at $k$ (denoted by Prec@$k$) as the evaluation metric of the prediction performance. Prec@$k$ denotes the number of correct labels in the top $k$ predictions. We run our code and all other baselines on a Linux machine with 40 cores and 128 GB RAM. We implemented our prediction Algorithms \ref{alg:pred} and \ref{alg:pred2} in \textsc{Matlab}. Learning Algorithms \ref{alg:PPMI} and \ref{alg:PPMI2}  are implemented parlty in Python and partly in \textsc{Matlab}. The source code will be made available later. We evaluate three models (a)~\exmldsone{} i.e. Algorithm \ref{alg:PPMI} based on fast \ppmi{} matrix factorization for learning label embeddings as described in Section \ref{subsec:labelvec}, (b) \exmldstwo{} based on optimizing the objective \eqref{eqn:sgnsmll} as described in section \ref{sec:wordvec}, using neural network~\cite{mikolov2013distributed}  (c) \exmldsthree{} i.e. Algorithm \ref{alg:PPMI2} based on combined learning objective \eqref{eqn:sgnsmllcomb}.

\paragraph{Compared methods.} We compare our algorithms with the following baselines.
\begin{enumerate}
\item \sleec{}~\cite{bhatia2015sparse}, which was shown to outperform all other embedding baselines on the benchmark datasets.
\item \leml{}~\cite{yu2014large}, an embedding based method. This method also facilitates incorporating label information (though not proposed in the original paper); we use the code given by the authors of \leml{} which uses item features\footnote{https://goo.gl/jdGbDPl}. We refer to the latter method that uses label correlations as \lemlimc{}.
\item \fastxml{}~\cite{prabhu2014fastxml}, a tree-based method. 
\item \pdsparse{}~\cite{yen2016pd}, recently proposed embedding based method 
\item \pfastrexml{}~\cite{jain2016extreme} is an extension of \fastxml{}; it was shown to outperform all other tree-based baselines on benchmark datasets.
\item \dismec{}~\cite{babbar2016dismec} is recently proposed scalable implementation of the \onevsall{} method.
\item \dxml{}~\cite{zhang2017deep} is a recent deep learning solution for multi-label learning
\item \onevsall{}~\cite{zhang2017deep} is traditional one vs all multi-label classifier
\end{enumerate}

We report all baseline results from the the extreme classification repository. \footnote{http://manikvarma.org/downloads/XC/XMLRepository.html}, where they have been curated; note that all the relevant research work use the same train-test split for benchmarking.

\begin{table*}[h!]
\small
\centering
\caption{Dataset statistics}
\label{tab:datasets}
\begin{tabular}{ccccc}
\hline
\textbf{Dataset} & \textbf{Feature} & \textbf{Label} & \textbf{Train} & \textbf{Test} \\
\hline
Bibtex~\cite{katakis2008multilabel} & 1836 &159& 4880  &2515\\
Delicious~\cite{tsoumakas2008effective} &500&  983&  12920 &3185\\
EURLex-4K~\cite{loza2008efficient}& 5000& 3993& 15539&  3809\\
rcv1v2~\cite{lewis2004rcv1}&47236&101&3000&3000\\
Delicious-200K~\cite{tsoumakas2008effective} & 782585& 205443& 196606& 100095\\
MediaMill~\cite{snoek2006challenge} & 120 & 101 & 30993 & 12914\\
\hline
\end{tabular}
\end{table*}

\paragraph{Hyperparameters.} We use the same embedding dimensionality, preserve the same number of nearest neighbors for learning embeddings as well as at prediction time, and the same number of data partitions used in \sleec{}~\cite{bhatia2015sparse} for our method~\exmldsone and ~\exmldstwo. For small datasets, we fix negative sample size to 15 and number of iterations to 35 during neural network training, tuned based on a separate validation set. For large datasets (4 and 5 in Table~\ref{tab:datasets}), we fix negative sample size to 2 and number of iterations to 5, tuned on a validation set. In \exmldsthree{}, the parameters (negative sampling) are set identical to \exmldsone{}. For  baselines, we either report results from the respective publications or used the best hyper-parameters reported by the authors in our experiments, as needed.

\textbf{Performance evaluation.} The performance of the compared methods are reported in Table~\ref{tab:precision}. Performances of the proposed methods \exmldsone{} and \exmldstwo{} are found to be similar in our experiments, as they optimize the same objective \ref{eqn:sgnsmll}; so we include only the results of \exmldsone{} in the Table. We see that the proposed methods achieve competitive prediction performance among the state-of-the-art embedding and tree-based approaches. In particular, note that on Medialmill and Delicious-200K datasets our method achieves the best performance.

\textbf{Training time.} 
Objective \ref{eqn:sgnsmll} can be trained using a neural network, as described in~\cite{mikolov2013distributed}. For training the neural network model, we give as input the $k$-nearest neighbor instance pairs for each training instance $i$, where the neighborhood is computed in the space of the label vectors $\y_i$. We use the Google \texttt{word2vec} code\footnote{https://code.google.com/archive/p/word2vec/} for training. We parallelize the training on 40 cores Linux machine for speed-up. Recall that we call this method~\exmldstwo{}. We compare the training time with our method~\exmldsone{}, which uses a fast matrix factorization approach for learning embeddings. Algorithm \ref{alg:PPMI} involves a single SVD as opposed to iterative SVP used by \sleec{} and therefore it is significantly faster. We present training time measurements in Table \ref{tab:small_results}. As anticipated, we observe that \exmldstwo{} which uses neural networks is slower than~\exmldsone{} (with 40 cores). Also, among the smaller datasets, \exmldsone{} trains 14x faster compared to \sleec on Bibtex dataset. In the large dataset, Delicious-200K, \exmldsone{} trains 5x faster than \sleec{}.

\begin {table*}[h!]
\small
\centering
\caption {Comparing training times (in seconds) of different methods}
\label{tab:small_results}
\begin{tabular}{ccccccc}
\hline
Method  & Bibtex & Delicious & Eurlex & Mediamill  & Delicious-200K \\
         \hline
$\exmldsone$ & \bf{23} & \bf{259}  & \bf{580.9}  &  \textbf{1200} & \bf{1937} \\
\hline
$\exmldstwo$  & 143.19 & 781.94  & 880.64  &  12000 & 13000 \\
\hline
$\sleec$ & 313 & 1351  & 4660  &  8912 & 10000  \\
\hline
\end{tabular}
\end{table*}

\begin{table*}[h!]
\tiny
\begin{center}
\caption {Comparing prediction performance of different methods($-$ mean unavailable results). Note that although SLEEC performs slightly better, our model is much faster as shown in the results in Table~\ref{tab:small_results}. Also note the performance of our model in Table~\ref{tab:ExMLDS3} when a significant fraction of labels are missing is considerably better than SLEEC}
\label{tab:precision}
\begin{tabular}{c|c|ccccc|cc|cc}
\hline
\multicolumn{1}{c}{Dataset} & \multicolumn{1}{|c|}{Prec@k}  &  \multicolumn{5}{c}{Embedding Based} & \multicolumn{2}{|c|}{Tree Based} & \multicolumn{2}{c}{Others}\\
         &  & \exmldsone{} & \dxml{} & \sleec{} & \leml{} & \pdsparse{} & \pfastrexml{} & \fastxml{} & \onevsall{} & \dismec{} \\
\hline
Bibtex & \begin{tabular}{@{}c@{}}P@1\\P@3\\P@5 \end{tabular}&
\begin{tabular}{@{}c@{}}63.38\\38.00\\27.64\end{tabular}& 
\begin{tabular}{@{}c@{}}63.69\\37.63\\27.71\end{tabular}&
\begin{tabular}{@{}c@{}}\textbf{65.29}\\\textbf{39.60}\\\textbf{28.63}\end{tabular}& 
\begin{tabular}{@{}c@{}}62.54\\38.41\\28.21\end{tabular}& 
\begin{tabular}{@{}c@{}}61.29\\35.82\\25.74\end{tabular}& 
\begin{tabular}{@{}c@{}}63.46\\39.22\\29.14\end{tabular}&
\begin{tabular}{@{}c@{}}63.42\\39.23\\28.86\end{tabular}&
\begin{tabular}{@{}c@{}}62.62\\39.09\\28.79\end{tabular}&
\begin{tabular}{@{}c@{}}- \\-\\ -\end{tabular}\\
\hline
Delicious & \begin{tabular}{@{}c@{}}P@1\\P@3\\P@5\end{tabular} & 
\begin{tabular}{@{}c@{}}67.94\\61.35\\56.3 \end{tabular}& 
\begin{tabular}{@{}c@{}}67.57\\61.15\\56.7\end{tabular}&
\begin{tabular}{@{}c@{}}\textbf{68.10}\\61.78\\57.34\end{tabular}&
\begin{tabular}{@{}c@{}}65.67\\60.55\\56.08\end{tabular}& \begin{tabular}{@{}c@{}}51.82\\44.18\\38.95\end{tabular}&
\begin{tabular}{@{}c@{}}67.13\\62.33\\58.62\end{tabular}
&\begin{tabular}{@{}c@{}}69.61\\\textbf{64.12}\\\textbf{59.27}\end{tabular}
&\begin{tabular}{@{}c@{}}65.01\\58.88\\53.28\end{tabular}&
\begin{tabular}{@{}c@{}}- \\-\\ -\end{tabular}\\
\hline
Eurlex & \begin{tabular}{@{}c@{}}P@1 \\ P@3\\ P@5 \end{tabular} & 
\begin{tabular}{@{}c@{}}77.55\\64.18\\52.51 \end{tabular}& 
\begin{tabular}{@{}c@{}}77.13\\64.21\\52.31 \end{tabular}&
\begin{tabular}{@{}c@{}}79.52\\64.27\\52.32\end{tabular}& 
\begin{tabular}{@{}c@{}}63.40\\50.35\\41.28\end{tabular}& \begin{tabular}{@{}c@{}}76.43\\60.37\\49.72\end{tabular}&
\begin{tabular}{@{}c@{}}75.45\\62.70\\52.51\end{tabular}&
\begin{tabular}{@{}c@{}}71.36\\59.90\\50.39\end{tabular}&
\begin{tabular}{@{}c@{}}79.89\\66.01\\53.80\end{tabular}&
\begin{tabular}{@{}c@{}} \textbf{82.40} \\ \textbf{68.50 }\\ \textbf{57.70}\end{tabular}\\
\hline
Mediamill & \begin{tabular}{@{}c@{}}P@1 \\ P@3\\ P@5 \end{tabular} & 
\begin{tabular}{@{}c@{}}87.49\\\textbf{72.62}\\\textbf{58.46}\end{tabular}&
\begin{tabular}{@{}c@{}}\textbf{88.71}\\71.65\\56.81\end{tabular}&
\begin{tabular}{@{}c@{}}87.37\\72.6\\58.39\end{tabular}& \begin{tabular}{@{}c@{}}84.01\\67.20\\52.80\end{tabular}& \begin{tabular}{@{}c@{}}81.86\\62.52\\45.11\end{tabular}&
\begin{tabular}{@{}c@{}}83.98\\67.37\\53.02\end{tabular}&
\begin{tabular}{@{}c@{}}84.22\\67.33\\53.04\end{tabular}&
\begin{tabular}{@{}c@{}}83.57\\65.60\\48.57\end{tabular}&
\begin{tabular}{@{}c@{}}- \\-\\ -\end{tabular}\\
\hline
Delicious-200K & \begin{tabular}{@{}c@{}}P@1 \\ P@3\\ P@5 \end{tabular} & 
\begin{tabular}{@{}c@{}}46.07\\41.15\\38.57\end{tabular}&
\begin{tabular}{@{}c@{}}44.13\\39.88\\37.20\end{tabular}&\begin{tabular}{@{}c@{}}\textbf{47.50}\\\textbf{42.00}\\\textbf{39.20}\end{tabular}& \begin{tabular}{@{}c@{}}40.73\\37.71\\35.84\end{tabular}& \begin{tabular}{@{}c@{}}34.37\\29.48\\27.04\end{tabular}&
\begin{tabular}{@{}c@{}}41.72\\37.83\\35.58\end{tabular}&
\begin{tabular}{@{}c@{}}43.07\\38.66\\36.19\end{tabular}&
\begin{tabular}{@{}c@{}} -\\-\\-\end{tabular}&
\begin{tabular}{@{}c@{}} 45.50 \\38.70\\ 35.50\end{tabular}\\
\hline
\end{tabular}
\end{center}
\end{table*}

\textbf{Coping with missing labels.} In many real-world scenarios, data is plagued with lots of missing labels. A desirable property of multi-label learning methods is to cope with missing labels, and yield good prediction performance with very few training labels. In the dearth of training labels, auxiliary information such as label correlations can come in handy. As described in Section \label{subsec:labelcorr}, our method~\exmldsthree{} can learn from additional information. The benchmark datasets, however, do not come with auxiliary information. To simulate this setting, we hide 80\% non-zero entries of the training label matrix, and reveal the 20\% training labels to learning algorithms. As a proxy for label correlations matrix $C$, we simply use the label-label co-occurrence from the 100\% training data, i.e. $C$ = $Y^TY$ where $Y$ denotes the full training matrix. We give higher weight $\mu_1$ to $\mathbb{O}^1$ during training in Algorithm \ref{alg:PPMI2}. For prediction, We use Algorithm \ref{alg:pred2} which takes missing labels into account. We compare the performance of ~\exmldsthree with \sleec, \leml{} and \lemlimc in Table \ref{tab:ExMLDS3}. Note that while \sleec{} and \leml{} methods do not incorporate such auxiliary information, \lemlimc{} does. In particular, we use the spectral embedding based features i.e. SVD of $YY^T$ and take all the singular vectors corresponding to non-zero singular values as label features. It can be observed that on all three datasets, \exmldsthree{}~performs significantly better by huge margins. In particular, the lift over~\lemlimc{} is significant, even though both the methods use the same information. This serves to demonstrate the strength of our approach.

\begin{table}[H]
\small
\centering
\caption {Evaluating competitive methods in the setting where 80\% of the training labels are hidden}
\label{tab:ExMLDS3}
\begin{tabular}{cccccc}
\hline
Dataset         & Prec@k & \exmldsthree & \sleec & \leml & \lemlimc \\
\hline
Bibtex & \begin{tabular}{@{}c@{}}P@1 \\ P@3\\ P@5 \end{tabular} & \begin{tabular}{@{}c@{}}\textbf{48.51} \\ \textbf{28.43}\\ \textbf{20.7} \end{tabular} & \begin{tabular}{@{}c@{}}30.5 \\ 14.9\\ 9.81\end{tabular}& \begin{tabular}{@{}c@{}}35.98 \\ 21.02\\ 15.50\end{tabular}& \begin{tabular}{@{}c@{}}41.23 \\ 25.25\\ 18.56\end{tabular}\\
\hline
Eurlex & \begin{tabular}{@{}c@{}}P@1 \\ P@3\\ P@5 \end{tabular} & \begin{tabular}{@{}c@{}}\textbf{60.28}\\\textbf{44.87}\\\textbf{35.31} \end{tabular} & \begin{tabular}{@{}c@{}}51.4\\37.64\\29.62\end{tabular}&\begin{tabular}{@{}c@{}}26.22\\22.94\\19.02\end{tabular}& \begin{tabular}{@{}c@{}}39.24\\32.66\\26.54\end{tabular}
\\
\hline
rcv1v2 & \begin{tabular}{@{}c@{}}P@1 \\ P@3\\ P@5 \end{tabular} & \begin{tabular}{@{}c@{}}\textbf{81.67}\\\textbf{52.82}\\\textbf{37.74} \end{tabular} & \begin{tabular}{@{}c@{}}41.8\\17.48
\\10.63\end{tabular}&\begin{tabular}{@{}c@{}}64.83\\42.56
\\31.68\end{tabular}& \begin{tabular}{@{}c@{}}73.68\\48.56
\\34.82\end{tabular}
\\
\hline
\end{tabular}
\end{table}

\cleardoublepage
\paragraph{Joint Embedding and Regression.} 
We extended SGNS objective for joint training i.e. learning of embeddings $Z$ and regressor $V$ simultaneously.

\[
{\mathbb{O}_{i}^{t+1}} = {\mathbb{O}_{i}^{t}} + \eta{ \nabla_{V} {\mathbb{O}_{i}}}
\]

Gradient of objective \ref{eqn:sgnsmll} w.r.t to $V$ i.e. $\nabla_{V} {\mathbb{O}_{i}}$ is describe in detail below :

Given,

\[
K_{ij} = \langle \z_{i}\z_{j} \rangle = \langle {\z_{i}^T} \z_{j} \rangle
\]

\[
\z_{i} = V\x_{i}, \text{where } \text{V} \in \mathbb{R}^{d' \times d} 
\]

Objective \ref{eqn:sgnsmll} :
\begin{equation}
\begin{split}
 \max_{\z_1,\z_2,\dots,\z_n} \sum_{i=1}^n \bigg( {\sum_{j: \mathcal{N}_k(\y_i)}} \log\big(\sigma( \langle \z_{i}, \z_{j} \rangle)\big) + \\ \frac{n_-}{n} \sum_{j'} \log\big(\sigma(-  \langle \z_{i}, \z_{j'} \rangle )\big)\bigg),   
 \label{eqn:sgnsjointlearning}
 \end{split}
\end{equation}

rewriting with s.t.t $V$ and $K_{ij}$, we obtained

\begin{equation}
\begin{split}
\max_{V} \sum_{i=1}^n \bigg( {\sum_{j: \mathcal{N}_k(\y_i)}} \log\big(\sigma( \langle V\x_{i}, V\x_{j} \rangle)\big) + \\ \frac{n_-}{n} \sum_{j'} \log\big(\sigma(-  \langle V\x_{i}, V\x_{j'} \rangle )\big)\bigg),
\end{split}
\end{equation}
 
\[
 \max_{V} \sum_{i=1}^n \bigg( {\sum_{j: \mathcal{N}_k(\y_i)}} \log\big(\sigma( K_{ij} )\big) + \frac{n_-}{n} \sum_{j'} \log\big(\sigma(-  K_{ij'} )\big)\bigg),   
\]

rewriting for only $i^{th}$ instance, we have
\[
   {\mathbb{O}_{i}} =  {\sum_{j: \mathcal{N}_k(\y_i)}} \log\big(\sigma( K_{ij} )\big) + \frac{n_-}{n} \sum_{j'} \log\big(\sigma(-  K_{ij'} )\big),   
\]

\[
    \nabla_{V} {\mathbb{O}_{i}} =  {\sum_{j: \mathcal{N}_k(\y_i)}} \sigma(- K_{ij})\nabla_{V} {K_{ij}} - \frac{n_-}{n} \sum_{j'} \sigma( K_{ij'})\nabla_{V} {K_{ij'}}
\]

here, $\nabla_{V} {K_{ij}}$ can be obtain through,
\[
\nabla_{V} {K_{ij}} = V(\x_{i}\x_{j}^T + \x_{j}\x_{i}^T) = \z_i.\x_j^T + \z_j.\x_i^T
\]

Sometime cosine similarity perform better then dot product because of scale invariant, in that case the gradient would modify to :

\begin{equation}
\begin{split}
K_{ij} = \langle \frac{\z_{i}}{\|\z_{i}\|}\frac{\z_{j}}{\|\z_{j}\|} \rangle = \frac{ \langle{\z_{i}^T}\z_{j}\rangle}{\|\z_{i}\|\|\z_{j}\|}
\end{split}
\end{equation}

\begin{equation}
\begin{split}
\nabla_{V} {\langle\z_{i}\z_{j}\rangle} = \nabla_{V}(V\x_i)\z_j + \nabla_{V}(V\x_j)\z_i \\ = \langle\z_i\x_j^T\rangle + \langle \z_j\x_i^T\rangle = V(\langle\x_i\x_j^T + \x_j\x_i^T\rangle)
\end{split}
\end{equation}

\begin{equation}
\begin{split}
\nabla_{V} {\frac{1}{\|\z_{i}\|}} = \nabla_{V} {\z_{i}\z_{i}^T}^{\frac{-1}{2}} = \frac{-1}{2} {\z_{i}\z_{i}^T}^{\frac{-3}{2}}\nabla_{V} {\z_{i}\z_{i}^T} =\\ \frac{-1}{2} {\z_{i}\z_{i}^T}^{\frac{-3}{2}}{\x_{i}\z_{i}^T}
\end{split}
\end{equation}
\\
\begin{equation}
\begin{split}
\nabla_{V} {\frac{1}{\|\z_{j}\|}} = \nabla_{V} {\z_{j}\z_{j}^T}^{\frac{-1}{2}} = \frac{-1}{2} {\z_{j}\z_{j}^T}^{\frac{-3}{2}}\nabla_{V} {\z_{j}\z_{j}^T} =\\ \frac{-1}{2} {\z_{j}\z_{j}^T}^{\frac{-3}{2}}{\x_{j}\z_{j}^T}
\end{split}
\end{equation}
\\ \\
Let,
\[
a = \z_i^T\z_j , b = \frac{1}{\|\z_{i}\|} , c = \frac{1}{\|\z_{j}\|}
\]

\[
\nabla_{V} {K_{ij}} = -ab^3c\z_i(\x_{i})^T -abc^3\z_j(\x_{j})^T + bc(\z_i\x_j^T + \z_j\x_i^T)
\]

Gradient update after $t^{th}$ iteration for $i^{th}$ instance,

\[
{\mathbb{O}_{i}^{t+1}} = {\mathbb{O}_{i}^{t}} + \eta{ \nabla_{V} {\mathbb{O}_{i}}}
\]

\begin{table*}[ht!]
\centering
\caption {Performance on multiple large datasets with Joint Learning}
\label{tab:ExMLDS3}
\begin{tabular}{|c|c|c|c|c|c|c|}
\hline
& AmazonCat-13K & Wiki10K-31K & Delicious-200K & WikiLSHTC-325K & Wikipedia-500K & Amazon-670K \\ \hline
Prec@1 & 93.05 & 86.82 & 47.70 & 62.15 & 62.27 & 41.47 \\ \hline
Prec@2 & 86.56 & 80.44 & 43.67 & 48.37 & 49.60 & 38.58 \\ \hline
Prec@3 & 79.18 & 74.30 & 41.22 & 39.58 & 41.43 & 36.35\\ \hline
Prec@4 & 72.06 & 68.61 & 39.37 & 33.52 & 35.68 & 34.20 \\ \hline
Prec@5 & 64.54 & 63.68 & 37.98 & 29.10 & 31.42 & 32.43\\ \hline
nDCG@1 & 93.05 & 86.82 & 47.70 & 62.15 & 62.27 & 41.47\\ \hline
nDCG@2 & 89.60 & 81.89 & 44.63 & 57.00 & 55.23 & 39.73\\ \hline
nDCG@3 & 87.72 & 77.22 & 42.75 & 55.20 & 52.11 & 38.41\\ \hline
nDCG@4 & 86.35 & 72.92 & 41.34 & 54.77 & 50.59 & 37.31 \\ \hline
nDCG@5 & 85.92 & 69.13 & 40.27 & 54.84 & 49.88 & 36.46 \\ \hline
\end{tabular}
\label{table:jointlearn}
\end{table*}

To implement joint learning, we modified the existing public code of state of art embedding based extreme classification approach AnnexML ~\cite{Tagami:2017:AAN:3097983.3097987} \footnote{Code: https://research-lab.yahoo.co.jp/en/software/}, by replacing the \texttt{DSSM} \footnote{https://www.microsoft.com/en-us/research/project/dssm/} training objective by \texttt{word2vec} objective, while keeping cosine similarity, partitioning algorithm, and approximate nearest prediction algorithm same. For efficient training of rare label, we keep the coefficient ratio of negative to positive samples as \textit{20:1}, while training. We used the same hyper-parameters i.e.embedding size as 50, number of learner for each cluster as 15,  number of nearest neighbor as 10, number of embedding and partitioning iteration both 100, gamma as 1, label normalization as true, number of threads as 32. We obtain state of art result i.e. similar (some dataset slightly better also)  to {}\dismec ~\cite{babbar2016dismec}, {}\ppdsparse ~\cite{Yen:2017:PPP:3097983.3098083} and {}\annexml ~\cite{Tagami:2017:LEM:3041021.3054204} on all large datasets, see table \ref{table:jointlearn} for details results.

\section{Conclusions and Future Work}
We proposed a novel objective for learning label embeddings for multi-label classification, that leverages \texttt{word2vec} embedding technique; furthermore, the proposed formulation can be optimized efficiently by ~\sppmi{} matrix factorization. Through comprehensive experiments, we showed that the proposed method is competitive compared to state-of-the-art multi-label learning methods in terms of prediction accuracies. We also extended SGNS objective for joint learning of embeddings $Z$ and regressor $V$ and obtain state of art results. We proposed a novel objective that incorporates side information, that is particularly effective in handling missing labels.

\bibliographystyle{aaai}
\bibliography{aaai} 
\normalsize

\cleardoublepage
\appendix

\begin{center}
\Large\textbf{Leveraging Distributional Semantics for \\ Multi-Label Learning}
\end{center}

\subsection{A2. SGNS Objective as Implicit SPPMI factorization}
The SGNS~\cite{mikolov2013distributed} objective is as follows:

\[
 {\mathbb{O}_{i}} = {\sum_{j\in S_i}} {\log(\sigma( K_{ij}))} + {\sum_{k \sim P_{D}}^{M}}{\mathbb{E}_{k \sim P_{D}}} [\log(\sigma(-  K_{ik}))]   
\] 

 where, $P_{D}$ = $\frac{(\#k)^{0.75}}{\#D} $, $D$ is collection of all word-context pairs and $K_{ij}$ represent dot-product similarity between the embeddings of a given word (i) and context (j). 
 
 Here, $\#k$ represent total number of word-context pairs with context (k).
 
\[
 {\mathbb{O}_{\{i,j\}}} = \log(\sigma( K_{ij})) + {\sum_{k \sim P_{D}}^{\frac{M}{|S|}}}{\mathbb{E}_{k \sim P_{D}}} [\log(\sigma(-  K_{ik}))]   
\]

\[
{\mathbb{E}_{k \sim P_{D}}} [\log(\sigma(- K_{ik}))] = \sum_{k \sim P_{D}} \frac{(\#k)^{0.75}}{\#D} \log(\sigma(- K_{ik}))
\]

\[
{\mathbb{E}_{k \sim P_{D}}} [\log(\sigma(- K_{ik}))] = \frac{(\#j)^{0.75}}{\#D} \log(\sigma(- K_{ij})) \]\[ +
 \sum_{k \sim P_{D} \& k \neq j} \frac{(\#k)^{0.75}}{\#D} \log(\sigma(- K_{ik}))
\]
\\
Therefore,\\
\[
{\mathbb{E}_{j \sim P_{D}}} [\log(\sigma( - K_{ij}))] =  \frac{(\#j)^{0.75}}{\#D} \log(\sigma( - K_{ij}))
\]

\[
 {\mathbb{O}_{\{i,j\}}} = \log(\sigma( K_{ij})) +  \frac{M}{|S|} \frac{(\#j)^{0.75}}{\#D} \log(\sigma(- K_{ij}))   
\]
\\
Let $\gamma K_{ij}$ = x, then
\[
 \nabla_{x}\mathbb{O}_{\{i,j\}}  = \sigma(-x) - \frac{M}{|S|} \frac{(\#j)^{0.75}}{\#D}\sigma(x)
\]
equating $\nabla_{x}\mathbb{J}_{\{i,j\}}$ to 0, we get :

\[
e^{2x} -\left(\frac{1}{\frac{M}{|S|}\frac{(\#j)^{0.75}}{\#D}} -1 \right) e^x - \left(\frac{1}{\frac{M}{|S|}\frac{(\#j)^{0.75}}{\#D}}\right) = 0
\]
\\
If we define $y$ = $e^x$ , this equation becomes a quadratic equation of y, which has two solutions, $y$ =- $1$ (which is invalid given the definition of y) and
\\
\[
y = \frac{1}{\frac{M}{|S|}\frac{(\#j)^{0.75}}{\#D}} = \frac{\#D*|S|}{M*(\#j)^{0.75}}
\]

Substituting y with $e^x$ and x with $K_{ij}$ reveals :
\[
K_{ij} = \log\left(\frac{\#D*|S|}{M*(\#j)^{0.75}}\right)
\]
\\
Here $|S|$ = $\#(i,j)$ and $M$ = $\mu \#(i)$ i.e. $\mu$ proportion of total number of times label vector (i) appear with others.

\[
K_{ij} = \log\left(\frac{\#(i,j)(\#D)}{\#(i)(\#j)^{0.75}}\right) - log(\mu) 
\]

\[
K_{ij} = \log\left(\frac{P(i,j)}{P(i)P(j)}\right) - log(\mu) 
\]
Here P(i,j),P(i) and P(j) represent probability of co-occurrences of $\{i,j\}$ , occurrence of i and occurrence of j respectively,\\
Therefore,
\[
K_{ij}= \pmi_{ij} -\log(\mu) = \log(P(i|j)) - log(\mu)
\]

Note that $PMI^+$ is inconsistent, therefore we used the sparse and consistent positive \pmi (\ppmi) metric, in which all negative values and nan are replaced by 0:

\[
\ppmi_{ij} = \max(\pmi_{ij},0)
\]

Here, PMI is point wise mutual information and PPMI is positive point wise mutual information. Similarity of two $\{i,j\}$ is more influenced by the positive neighbor they share than by the negative neighbor they share as \textit{uninformative} i.e. 0 value. Hence, SGNS objective can be cast into  a weighted matrix factorization problem, seeking the optimal lower d-dimensional factorization of the matrix SPPMI under a metric which pays more for deviations on frequent $\#(i,j)$ pairs than deviations on infrequent ones. 

Using a similar derivation, it can be shown that noise-contrastive estimation (NCE) which is alternative to (SGNS)  can be cast as factorization of (shifted) log-conditional-probability matrix

\[
K_{ij} = \log\left(\frac{\#(i,j)}{(\#j)}\right) - \log(\mu)
\]


\end{document}